\documentclass[10pt, a4paper]{article}

\usepackage[final]{lrec2026} 
\usepackage{multirow}
\usepackage{gb4e}
\usepackage{natbib}

\title{EDDA-Coordinata: An Annotated Dataset of Historical Geographic Coordinates}

\name{Ludovic Moncla\textsuperscript{1}, Pierre Nugues\textsuperscript{2}, Thierry Joliveau\textsuperscript{3}, Katherine McDonough\textsuperscript{4}} 

\address{
    \textsuperscript{1}INSA Lyon, CNRS, Lyon 1 Universit\'e, LIRIS, UMR5205, 69621 Villeurbanne, France, \\
    \textsuperscript{2}Lund University, Lund Sweden \\
    \textsuperscript{3}Université de Saint-Etienne, CNRS, EVS UMR5600, Saint-Etienne, France \\
    \textsuperscript{4}Lancaster University, Lancaster, United Kingdom \\
         }

\abstract{
This paper introduces a dataset of enriched geographic coordinates retrieved from Diderot and d’Alembert's eighteenth-century \textit{Encyclopédie}. 
Automatically recovering geographic coordinates from historical texts is a complex task, as they are expressed in a variety of ways and with varying levels of precision. To improve retrieval of coordinates from similar digitized early modern texts, we have created a gold standard dataset, trained models, published the resulting inferred and normalized coordinate data, and experimented applying these models to new texts.
From 74,000 total articles in each of the digitized versions of the \textit{Encyclopédie} from ARTFL and ENCCRE, we examined 15,278 geographical entries, manually identifying 4,798 containing coordinates, and 10,480 with descriptive but non-numerical references.
Leveraging our gold standard annotations, we trained transformer-based models to retrieve and normalize coordinates. The pipeline presented here combines a classifier to identify coordinate-bearing entries and a second model for retrieval, tested across encoder–decoder and decoder architectures. Cross-validation yielded an 86\% EM score. 
On an out-of-domain eighteenth-century Trévoux dictionary (also in French), our fine-tuned model had an 61\% EM score, while for the nineteenth-century, 7th edition of the \textit{Encyclopædia Britannica} in English, the EM was 77\%.
These findings highlight the gold standard dataset’s usefulness as training data, and our two-step method’s cross-lingual, cross-domain generalizability. 
\\ \newline  }

\begin{document}

\newcommand{\comment}[1]{\textcolor{red}{\textbf{#1}}}

\maketitleabstract

\section{Introduction}

Geographic coordinates have been communicated in historical documents since there have been means for measuring the size and shape of the earth. In early modern Europe, the shift to communicating about location using coordinates rather than prose descriptions of well-known frontier areas, natural features like mountain ranges and rivers, cities, or other sites was not a linear process, despite the increasing quantification of the sciences \citep{edney1999}. Enlightenment encyclopedias, dictionaries, and other reference works used coordinates to complement descriptions, but they were rarely expressed in a standard format. Scientific communities did not adopt international standards for weights and measures until the nineteenth century \citep{quinn2012}, and so it is typical that geospatial coordinates styles were highly varied throughout the eighteenth and nineteenth centuries. 

To recover and spatially analyze coordinates from historical texts, we need a method for retrieving and normalizing coordinates, while preserving a link to their original expression and location in the text. The method we present here is a novel approach to automating coordinate retrieval and reformatting from unstructured, digitized texts. With such data, for the first time it will be possible to assess, across thousands of coordinates in early modern and modern reference works, patterns in communicating about coordinate precision, and the geospatial footprint of places referred to by numerical coordinates compared to descriptive text. 

Here, we present a gold standard dataset of coordinates collected from the major French Enlightenment project, the \textit{Encyclopédie} \citep{alembert1751}. Working with both digitized versions \citeplanguageresource{ARTFL2022, ENCCRE2017}, we independently identified relevant geographical entries and annotated each set.
Thus, from 74,000 \textit{Encyclopédie} entries, 15,278 describe places (cities, regions, rivers, etc.), where locations may be specified with coordinates or in prose descriptions. Two independent annotators manually annotated coordinates, yielding an agreement rate of 0.98 for points and 0.52 for surfaces. After reconciling the ARTFL and ENCCRE entries, we obtained a dataset of 4,798 entries with explicit coordinates, and 10,480 entries with only prose location information.  

We then trained transformer-based models to automate coordinate detection and normalization in two steps: first, we use a classifier to identify whether an entry contains coordinates; next, we apply a sequence-to-sequence model to retrieve and normalize them. We experimented with encoder–decoder and decoder-only architectures, and cross-validation yielded an 86\% EM score for coordinate retrieval. To assess robustness, we applied our best model to new texts. 

For this paper, we: 1) perform a double annotation of all geographic entries in EDDA, focusing on locations described with coordinates; 2) reconcile and merge these annotations into a single gold standard dataset, yielding 4,798 coordinate-bearing articles; 3) fine-tune a sequence-to-sequence model for coordinate retrieval and normalization;  
4) and apply and evaluate these models on external datasets.
The gold standard dataset, models, code, and a demonstration are available on HuggingFace\footnote{\url{https://huggingface.co/GEODE}} and github\footnote{\url{https://github.com/GEODE-project/edda-coordinata}}.

\section{Geographic Coordinates in the \textit{Encyclopédie}}
Geographic coordinates are expressed in the \textit{Encyclopédie} as either points or bounding boxes. Points correspond principally to towns and cities, while bounding boxes are associated with regions, islands, or countries. Below, we provide examples of these two different geometries. 

\subsection{Coordinate Types}

\paragraph{Single points.} Most coordinates correspond to a single point expressed in latitude and longitude. For example:

\begin{exe}
\ex * AAHUS, s. petite ville d’Allemagne dans le cercle de Westphalie, capitale de la Comté d’Aahus. \textbf{\textit{Long. 24. 36. lat. 52. 10.}}
\trans  \textit{* AAHUS, s. small city in Germany in the circle of Westphalia, capital of the County of Aahus. \textbf{\textit{Long. 24. 36. lat. 52. 10.}}}
\end{exe}
The precision can be degrees (D) only, degrees and minutes (DM), or degrees, minutes, and seconds (DMS). Sometimes, the location is given by only the latitude \textit{or} the longitude as in:

\begin{exe}
\ex * AGRIGNON, (Géog.) l’une des îles des Larrons ou Mariannes. \textbf{\textit{Lat. 19. 40.}}
\trans \textit{* AGRIGNON, (Geog.) one of the islands of Ladrones or Marianas. \textbf{\textit{Lat. 19. 40.}}}
\end{exe}

\paragraph{Bounding boxes.} In addition to points, some entries describe countries or regions with their maximal extensions in longitude and latitude. This corresponds to a rectangle. For example:
\begin{exe}
\ex * ABISSINIE, s. f. grand Pays \& Royaume d’Afrique. \textbf{\textit{Long. 48-65. lat. 6-20.}}
\trans  \textit{* ABYSSINIA, n.f. large Country and Kingdom of Africa. \textbf{\textit{Long. 48-65. lat. 6-20.}}}
\end{exe}
For single points, precision varies widely. Some regions are also bounded by latitudes or longitudes only.

\paragraph{Polygonal chains.} 
For some rivers, there are instances of connected points. They consist mostly of two points designating the source and mouth, but sometimes include intermediate points. Points in a connected sequence can be incomplete, missing either latitude or longitude. For example:
\begin{exe}
\ex * AMUR ou AMOER, riviere de la grande Tartarie en Asie ; elle a sa source près du lac Baycal, vers le \textbf{\textit{117. degré de longitude}}, \& se jette dans l’Océan oriental au \textbf{\textit{55. degré de latitude septentrionale, \& le 152. de longitude}}... 
\trans * AMUR or AMOER, river of Great Tartary in Asia; its source is near lake Baikal, around \textbf{\textit{117. degree of longitude}}, \& and it flows into the oriental Ocean at \textbf{\textit{55. degree north latitude, \& 152. of longitude}}...
\end{exe}

\subsection{Sequences}
A few entries contain two or more coordinate instances of the previous types. These sequences correspond to two different cases: subentries or multiple sources.

\paragraph{Subentries.} Some entries have subentries containing coordinates. \textit{Ava}, for example, has three subentries describing different kingdoms. These are shown below in different colors, each documented with the coordinates of their capitals.
\begin{exe}
\ex * AVA, (\textit{Géog. mod.}) royaume d’Asie, sur la riviere de même nom, au-delà du Gange, sur le golfe de Bengale. \textit{Ava} en est la capitale~; sa \textbf{\textit{longitude} est 114, \& sa \textit{latit. 21.}} \textcolor{blue}{Il y a au Japon un royaume du même nom, dont la capitale s’appelle aussi \textit{Ava}~: ce royaume est renfermé dans une île [...]. \textbf{\textit{long. 151, 10, lat. 33.}}} \textcolor{purple}{\textit{Ava}, autre royaume du Japon, avec une ville de même nom, dans la presqu’île de Niphon. \textbf{\textit{Long. 159, lat. 35, 20.}}}
\trans  \textit{* AVA, (Geog. mod.) kingdom of Asia, on the river of the same name, beyond the Ganges, on the Bay of Bengal. \textit{Ava} is the capital; its \textbf{\textit{longitude} is 114, and its \textit{latit. 21.}} \textcolor{blue}{There is in Japan a kingdom of the same name, whose capital is also called \textit{Ava:} this kingdom is contained within an island [...]. \textbf{\textit{long. 151, 10, lat. 33.}}} \textcolor{purple}{\textit{Ava}, another kingdom of Japan, with a city of the same name, in the Niphon peninsula. \textbf{\textit{Long. 159, lat. 35, 20.}}}}
\end{exe}

\paragraph{Multiple sources.} Some entries cite multiple sources (e.g. other publications reporting this information) and values for the coordinates of one place. For example, the city of Autan-Keluran was described by Ulugh Beg and Nasir al-Din (al-Tusi) and the corresponding entry cites both:
\begin{exe}
\ex * AUTAN-KELURAN, (Géog.) ville du Turquestan. \textbf{\textit{Long. 110d. \& lat. 46. 45.}} selon Uluhbeg ; \& \textbf{\textit{long. 116. \& lat. 45.}} selon Nassiredden.
\trans  \textit{* AUTAN-KELURAN, (Geog.) city of Turkestan. \textbf{\textit{Long. 110d. \& lat. 46. 45.}} according to Uluhbeg; \& \textbf{\textit{long. 116. \& lat. 45.}} according to Nassiredden.}
\end{exe}

\subsection{Prime Meridians}

\textit{Encyclopédie} longitudes primarily use El Hierro, or the Meridian Island, as an implicit reference meridian. This convention was adopted in France and then throughout Europe in the early seventeenth century \citep{lagarde}. 
It is in fact a proxy for Paris meridian, which serves as the real reference. Faced with the difficulty of locating the Meridian Island precisely, in 1720 the cartographer Delisle arbitrarily set it at 20° west of Paris.\footnote{See the \textit{Encyclopédie} entry ``Méridien (géographie)'' for further discussion.} Nonetheless, a few entries use Paris, London, or Beijing as in:

\begin{exe}
\ex FONING, (Géog.) cité de la Chine dans la province de Fokien. \textbf{\textit{Long. 4. 0. latit. 26. 33.}} suivant le P. Martini qui place le \textbf{premier méridien au palais de Peking}.
\trans  \textit{FONING, (Geog.) city of China in the province of Fokien. \textbf{\textit{Long. 4. 0. latit. 26. 33.}} according to Father Martini who places the \textbf{prime meridian at Peking [Beijing] palace}.}
\end{exe}

\subsection{Converting Coordinates to Modern Standards}
In the \textit{Encyclopédie}, longitude coordinates are mostly expressed from 0° to 360° eastwards. We reformat these according to modern usage, from -180° westwards to +180° eastwards. They must also be expressed in reference to the Greenwich meridian and therefore reduced by -20° (the arbitrary difference between Meridian Island and Paris) and increased by +2° 20$'$ 14.025$''$ (the difference between Paris and Greenwich). In practice, we subtract -17.66° from the longitude coordinates converted to decimal degrees.\footnote{In the very rare cases where other meridians are mentioned, the values should be recalculated by setting a reference point for the original meridian found in the \textit{Encyclopédie}. We did not perform this calculation. Finally, southern latitudes are often omitted, which can lead to location errors.}

\section{Dataset Structure \& Format}
\label{sec:dataset}
We represent coordinates as strings and use a data structure based on nested lists to reflect all cases we observed in the corpus: points, surfaces, polygonal chains, and sequences. The strings enable us to remain as true as possible to the original text. Nested lists were sufficient to represent all examples.

\paragraph{Point coordinates.} Geographic point coordinates, either for single points or in rectangles, follow the latitude and longitude convention in degrees, minutes, and seconds, with the cardinal points, north or south, for latitudes and east or west for longitudes, for example: \verb=48 51' 20" N 20 21' 30" E=. We then placed these coordinates in strings: \verb="48 51' 20\" N 20 21' 30\" E"=. 

We made sure these strings corresponded to well-formed geographical points by creating coordinate objects from them using the \verb=geopy.Point()= class as for instance with:
\begin{verbatim}
from geopy import Point

Point("48 51' 20\" N 20 21' 30\" E")
\end{verbatim}
returning 
\begin{verbatim}
Point(48.85555555, 20.35833333, 0.0)
\end{verbatim}

\paragraph{Points and rectangles.}
We represent a point as a singleton and a rectangle as a pair, using lists. The length of a list therefore indicates the nature of the object it encodes. 

For homogeneous representation with sequences, we enclose the single points and rectangles in a list so that all lists have a depth of 2. We add the \verb=pchain= prefix to differentiate polygonal chains from rectangles:
\begin{itemize}
\item Point, \verb=[[str]]=, for example
\begin{itemize}
\item AAHUS \verb=[["52 10' N 24 36' E"]]=
\end{itemize}
\item  Rectangle \verb=[[str, str]]=:
\begin{itemize}
\item ABISSINIE \texttt{[['6 N 48 E', '20 N 65 E']]};
\item FALSTER \texttt{[["55 50' N 28 50' E", "56 50' N 29 26' E"]]}.
\end{itemize}
\item  Polygonal chains \texttt{[['pchain'] [[str], [str], ...]]}:
\begin{itemize}
\item AMUR ou AMOER \texttt{[['pchain'], ['117 E'], ['55 N 152 E']]}
\end{itemize}
\end{itemize}

\begin{table*}[tb]
\centering
\resizebox{\textwidth}{!}{
\begin{tabular}{llp{11cm}}
\hline
\textbf{Type}&\textbf{Headword}&\textbf{Coordinates}\\
\hline
Point&AAHUS &\texttt{[["52 10' N 24 36' E"]]}\\
Point& AGRIGNON&\texttt{[["19 40' N"]]}\\
Rectangle&ABISSINIE &\texttt{[['6 N 48 E', '20 N 65 E']]}\\
Rectangle&FALSTER &\texttt{[["55 50' N 28 50' E", "56 50' N 29 26' E"]]}\\
Polygonal chain&AMUR ou AMOER&\texttt{[['pchain'], ['117 E'], ['55 N 152 E']]}\\
Subentries&AVA&\texttt{[['subart'], ['21 N 114 E'], ["33 N 151 10' E"], ["35 20' N 159 E"]]}\\
Mult. sources&HEGETMATIA&\texttt{[['multsrc'], ['50 N 39 40\textbackslash' 11" E'], ["51 55' N 33 50' E"]]}\\
\hline
\end{tabular}
}
\caption{Annotation examples for points, rectangles, polygonal chains, and sequences.}
    \label{tab:examples}
\end{table*}

\paragraph{Sequences.} To store sequences, we use lists consisting of a specific prefix to identify the 2 possible cases followed by either points or rectangles:
\begin{enumerate}
    \item \verb=subart=, for a sequence of subentries, for instance the AVA entry:\\ 
\texttt{[['subart'], ['21 N 114 E'], ["33 N 151 10' E"], ["35 20' N 159 E"]]}
    \item \verb=multsrc=, for multiple sources, for instance HEGETMATIA:\\
 \texttt{[['multsrc'], ['50 N 39 40\textbackslash' 11" E'], ["51 55' N 33 50' E"]]}.
\end{enumerate}
We reserved a fourth prefix for unforseen cases, \verb=misc=.

\subsection{Dictionary Keys}
The dataset is a list of JSON dictionaries. Each dictionary represents 1 entry and has five keys: 
\begin{enumerate}
    \item unique identifier for entry;
    \item entry headword;
    \item entry text;
    \item list of coordinates with the key \verb='coordinates'=;
    \item list of Prime Meridians, if mentioned.
\end{enumerate}

When an entry uses a specific Prime Meridian: Paris, Beijing, London, or Lund, we add the \verb=meridian= key  as in FONING.
\begin{quote}
\texttt{'meridian': ['Pékin']}.
\end{quote}
In two cases, latitude is given relative to a reference point other than the equator. We indicate this in \verb=meridian=.

\section{Annotation Process}
\label{sec:annotation-process}
We independently annotated two digitized versions of the \textit{Encyclopédie}: the hand-keyed ARTFL project data \citep{morrissey1998encyclopedie} and the ENCCRE project data \citep{guilbaud2013entrer}. While the \textit{Encyclopédie}'s eighteenth-century editors classified many articles with categories such as Agriculture, Geography, History, Navigation, Medicine, and so on, these were applied in non-standard ways across the 17 volumes of text entries. The ARTFL and ENCCRE teams have sought to  normalize these classifications through both manual and automatic methods \citep{roe2016, horton2009}.\footnote{\url{https://enccre.academie-sciences.fr/encyclopedie/politique-editoriale/?s=23&}}  

To build our datasets, we first identified the articles describing a location using these categories:
\begin{enumerate}
\item For the ARTFL corpus, we developed a set of rules based on the occurrence of the keywords \textit{latitude} and \textit{longitude} (including their variant such as \textit{lat.}, \textit{long.}, \textit{latit.}, etc.) in combination with numerical expressions. These rules were formalized using the Corpus Query Language (CQL) and executed using TXM software\footnote{\url{https://txm.gitpages.huma-num.fr/textometrie/}}, which generated the corresponding concordances. After several iterations, we obtained a set of 4,458 articles.
\item For the ENCCRE corpus, we extracted and annotated all 15,274 articles labeled with the ENCCRE \textit{Geography} domain.
\end{enumerate}

\subsection{Consensus Identification}

Annotator 1 (a geographer) annotated coordinates in 4,650 entries from the ARTFL corpus. 4,505 were single points, of which 4,431 represent well-formed points with both latitude and longitude specified. 108 entry coordinates were annotated as surfaces, 28 as alternative coordinates or sequences, and 9 as coordinate pairs associated with linear objects (e.g., rivers).
Annotator 2 (a computer scientist) annotated 4,779 entries from the ENCCRE corpus. This included 4,508 single points (of which 4,272 are well-formed points), 136 surfaces, 84 entries with multiple sources, and 51 sequences of subentries.
Restricting the comparison to \textbf{well-formed single-point annotations}, the results are as follows:

\begin{itemize}
    \item The union of annotated entries is 4,382.
    \item 110 entries were annotated exclusively by Annotator 1, while 51 articles were annotated exclusively by Annotator 2.
    \item The intersection comprises 4,221 entries annotated by both annotators.
    \item Within the intersection, 4,140 entries contain identical coordinates, corresponding to an agreement rate of 0.981. For the 81 entries with divergent annotations, the calculated micro-average Character Error Rate (CER) is 0.185.
\end{itemize}

For \textbf{surface annotations}, the union of annotated articles amounts to 145, of which 52 were annotated exclusively by Annotator 1 and 5 by Annotator 2, leaving an intersection of 88 articles annotated by both. Within this intersection, 46 entries contain identical coordinates, corresponding to an agreement rate of 0.523. For the remaining 42 entries with divergent annotations, the calculated micro-average CER is 0.209. 

These results highlight the considerably lower level of agreement for surfaces compared to single-point annotations, reflecting the inherent difficulty of consistently delineating geographic extents and the greater interpretive variation this task entails. The low CER score indicates that divergent annotations remain closely aligned, with minor, single-character variations.

\subsection{Discrepancy Resolution}
To ensure the accuracy and reliability of the final dataset annotations, we conducted a systematic review of all articles where annotators disagreed. Each case was examined through a collaborative discussion, during which the annotators reconciled annotations and resolved discrepancies.

For well-formed single-point annotations, the reconciliation process involved evaluating the 81 divergent cases. Of these, 72 (88.9\%) were resolved by selecting the more accurate annotation from one of the two annotators. The remaining 9 cases (11.1\%) required the introduction of new corrections, as both initial annotations were deemed incorrect after reassessment.

For surface annotations, a similar approach was applied to the 42 divergent cases. Here, 36 (85.7\%) were resolved by adopting the more precise annotation from one of the two annotators, while 6 cases (14.3\%) necessitated corrections due to errors in both original annotations.
We repeated this methodology for all kinds of annotations.

This iterative process ensured that the final dataset consisted of consolidated annotations, each subjected to a triple-checked validation where necessary. 

\subsection{Dataset Composition}

The dataset contains 15,278 entries, of which 4,798 contain coordinate annotations. These are categorized into distinct spatial annotation types (see Section~\ref{sec:dataset}). Table~\ref{tab:stat} shows a detailed breakdown of the dataset's composition, including the distribution of annotation formats and precision levels. 
In the remaining 10,480 entries, the coordinates in the article did not describe the location of the headword. We also annotated 40 entries containing a Prime Meridian. 

\begin{table}[htb]
    \centering
    \begin{tabular}{llr}
    \hline
    \textbf{Category}          && \textbf{Count} \\ \hline
Simple types&    Well-formed points         & 4,287          \\
 &   Incomplete points          & 232            \\ 
  &  \hspace{1em} - Latitude only & 221          \\ 
&    \hspace{1em} - Longitude only & 11          \\ 
&    Surfaces                   & 133            \\ 
&    Polygonal chains        & 11             \\ 
\hline
Sequences&    Subentries                 & 47             \\ 
&    Multiple sources           & 87             \\ 
&    Miscellaneous              & 1              \\ \hline
    \end{tabular}
    \caption{Dataset statistics.}
    \label{tab:stat}
\end{table}

Table~\ref{tab:precisionmatrix} presents the distribution of coordinate precision formats for single well-formed points, where rows represent latitude formats and columns represent longitude formats. The precision of coordinates varies visibly across the dataset, with the most common format being Lat\_DM-Long\_DM (3,356 entries). Conversely, the least common formats suggest limited use of DMS for both latitude and longitude. Precision variation reflects differences in both the original sources and \textit{Encyclopédie} editorial practices.

\begin{table}[htb]
    \centering
    \begin{tabular}{lrrr}
    \hline
                     & Long\_D & Long\_DM & Long\_DMS \\ \hline
    Lat\_D  & 116             & 182              & 2                 \\ 
    Lat\_DM & 278             & 3,356            & 91                \\ 
    Lat\_DMS & 3              & 38               & 221               \\ \hline
    \end{tabular}
    \caption{Coordinate precision distribution for single well-formed points.}
    \label{tab:precisionmatrix}
\end{table}

\begin{figure*}[tb]
    \centering
    \includegraphics[width=.9\textwidth]{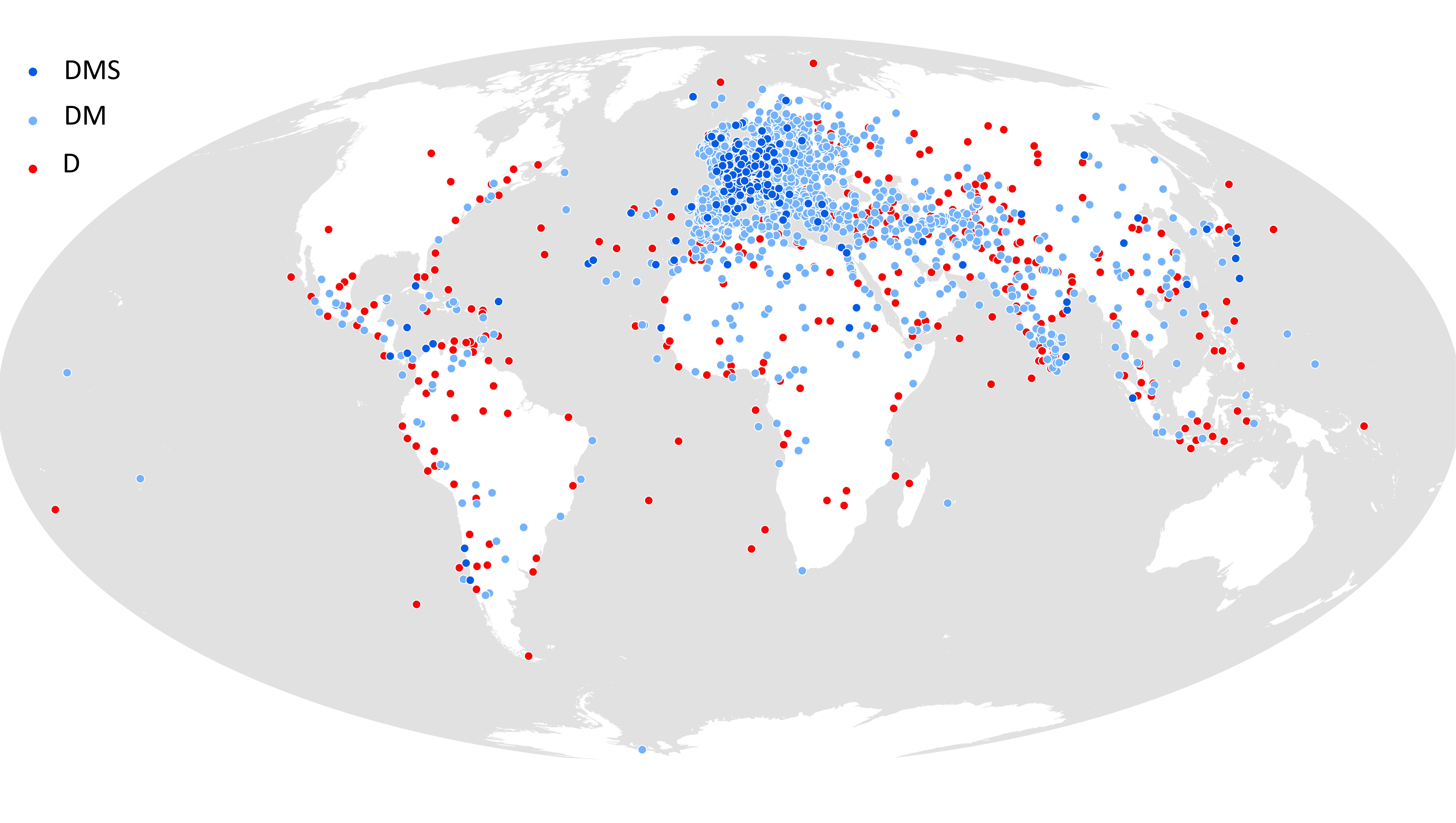}
    \caption{Map showing the location of dataset coordinates colored by their level of precision. Red points contain only degrees (D), light blue contain degrees and minutes (DM), and dark blue points contain degrees, minutes, and seconds (DMS).}
    \label{fig:precisionmap}
\end{figure*}

\section{Coordinate Precision}

Coordinate accuracy (in terms of being located in the correct place, based on knowledge available at the time) and precision (in terms of the level of detail in which the coordinate is expressed) varies widely across \textit{Encyclopédie} entries.\footnote{The further issue of whether a precise and accurate location in the eighteenth century is the same as a precise and accurate location in the twenty-first century is a separate, thorny challenge that we do not address here.} In the eighteenth century, less precise coordinates mean that some are inherently not accurate, but accuracy can also be a relic of incorrect source material. \citet{joliveau2024digital} have argued that precision varies as a function of the location's distance from Europe. Here, we refine our understanding of how \textit{Encyclopédie} editors and authors made use of more or less precise coordinates. We selected 3,693 entries that have the same level of measurement in latitude and longitude, i.e., both are expressed in only D (degrees), DM (degrees and minutes), and, the most precise, DMS (degrees, minutes, and seconds). All entries where latitude and longitude did not have the same level of precision (latitude was DM and longitude was only D) were excluded.
Measurements with M-level accuracy dominate (91\%), with 3\% at D-level and 7\% at DMS-level accuracy. As Figure~\ref{fig:precisionmap} shows, points expressed in DMS are mainly located in France, those expressed in DM are mainly in Europe and Asia, while the least precise points, in D only, are located in areas furthest from France.

By attaching each point to the nearest modern continent and country, it is possible to estimate the proportion of measurement levels for different modern geographical entities.
Each point (from the 3,693 entries) is linked to a modern country and continent (using Natural Earth and ESRI France data). Afterwards, we estimate the proportion of different levels of precision for the points grouped by modern countries and continents.
In these composite, modern groupings, point coordinates of any precision in Europe dominate over other continents. Europe has more than 3,000, while Asia has 422, Africa 161, and the Americas combined 92. 
The most common format of expressing coordinates is at the DM level (more than 75\% of total), but this varies by continent: 76\% in North America, 86\% in Asia, 89\% in Africa, 92\% in Europe, and 59\% in South America). 
The Americas are characterised by higher proportions of measurements at the D-level only (North America at 20\% and South America at 34\%). 
The proportions of DMS-level accuracy are much lower in Asia (1\%) and Africa (2\%) than in Europe (7\%) and, surprisingly, South America (also 7\%). France stands out at the European level with 22\% of its points at the DMS level, 78\% at the DM level, and almost none at the D (only) level. Major modern European countries are all more than 90\% at the DM level and less than 3\% at the DMS level, except Portugal, which is 7\% at the DMS level. Notably, France accounts for three-quarters of all DMS-level measurements (likely due to authors' access to recent tertiary surveying data and geographical dictionaries). Other high-precision coordinates are linked to specific sources used repeatedly across the volumes.
While a simple model linking coordinate precision levels to distance from France largely holds true, the exceptions to this model raised here require further, historical research.

\section{Model Training}

To evaluate the performance of different model architectures trained with our gold standard dataset, we apply this coordinate classification, retrieval, and normalization task on the \textit{Encyclopédie}. Last, we test these models on out-of-domain texts.

\subsection{Classification}
\label{sec:classification}
First, we automatically determine whether an article contains geographical coordinates. 
Many entries in the \textit{Encyclopédie} describe geographical entities without using numerical coordinate values, therefore only a subset is relevant for coordinate retrieval. 
We structure this as a binary classification task. Each article is represented by its raw text, truncated to 512 tokens to fit transformer input constraints. We fine-tune a pre-trained BERT \citep{Devlin2019} (multilingual cased) model with a binary classification head.
Our 4,798 annotated entries were used as positive examples, while the negative examples correspond to the other entries in the 15,278 ENCCRE ``Geography'' domain articles. 

We evaluate the classifier using a five-fold cross-validation protocol over four training epochs. We obtained a mean performance across folds of 99.2\% for the accuracy, 98.8\% for the precision, 98.6\% for the recall, corresponding to a F1 score of 98.7\%. The model achieved very high accuracy, precision, and recall, confirming that lexical patterns (e.g., recurrent use of abbreviations such as \textit{lat.} and \textit{long.}) are highly predictive of coordinate presence.
High precision and recall scores indicate that the classifier is highly robust, with few false positives or false negatives (note that \textit{precision} here refers to model results, not \textit{coordinate precision}).

\subsection{Retrieval and DMS Normalization}
Next, we retrieve coordinates and normalize their formats. This task presents significant challenges due to historical variations in notation, irregular use of abbreviations and symbols (e.g., ``d.'' for degrees, periods instead of colons, or missing cardinal directions), and multiple sources or alternative measurements in the same entry.

We frame coordinate retrieval as a sequence-to-sequence generation problem. We fine-tune \texttt{mt5-small}, a multilingual transformer-based encoder-decoder model, to generate normalized, DMS coordinate strings. The model was trained using the raw article text (still 512 tokens) as input, with the target output being one or more normalized coordinates concatenated into a single string. The dataset includes examples of both point coordinates and bounding box regions, with multiple sources encoded as described in Section~\ref{sec:dataset}.
The model was trained over 10 epochs using five-fold cross-validation to ensure robustness. 
We evaluate the model's performance on one fold (959 samples with 903 single well-formed points) using two metrics: Exact Match (EM) and CER. 

\begin{table}[ht]
    \centering
    \begin{tabular}{clccc}
               &             & EM            & CER  & Support \\ \hline
     &   \texttt{mt5-small}  & 0.86          & 0.07 & \multirow{2}{20pt}{959}\\ 
     &   \texttt{gpt5-mini}  & 0.86          & \textbf{0.03} \\ \hline
   points  &   \texttt{mt5-small}  & \textbf{0.92} & 0.01 & \multirow{2}{20pt}{903} \\ 
   only  &   \texttt{gpt5-mini}  & 0.90          & 0.01 \\ \hline
   others   &  \texttt{mt5-small}  & 0.00         & 0.57 & \multirow{2}{20pt}{56} \\ 
    only &   \texttt{gpt5-mini}  & \textbf{0.29}& \textbf{0.22} \\ \hline
    \end{tabular}
    \caption{Coordinate retrieval and DMS normalization model performance.}
    \label{tab:extraction-scores}
\end{table}

The results in Table~\ref{tab:extraction-scores} indicate that EM accuracy remains challenging (0.86 on average for the fine-tuned \texttt{mt5-small}), whereas the CER is substantially lower (0.07). This gap suggests that most discrepancies arise from minor formatting differences rather than substantive numerical errors.
The zero-shot performance of \texttt{gpt5-mini} is comparable to that of \texttt{mt5-small}, but achieves a lower CER (0.03).
When restricted to single, well-formed points, performance across models is similar. However, for less frequent coordinate types, performance declines for both models, with \texttt{gpt5-mini} consistently outperforming \texttt{mt5-small}.
These results suggest that larger pre-trained models generalize more effectively to historical coordinate formats, even without task-specific fine-tuning.

Table~\ref{tab:extraction-scores-precision} provides a detailed breakdown of EM scores across different precision coordinates for both the \texttt{mt5-small} and \texttt{gpt5-mini} models. The results highlight notable trends in model performance. The Degrees (D) and Degrees-Minutes (DM) formats are generally easier to identify than the Degrees-Minutes-Seconds (DMS) format. 
\texttt{gpt5-mini} model achieves higher EM scores for DMS level precision, indicating its robustness in managing complex or irregular notations and underrepresented formats (see Table~\ref{tab:precisionmatrix}). Our fine-tuned \texttt{mt5-small} outperforms \texttt{gpt5-mini} on the well represented D-DM and DM-DM formats.
However, results indicate that the heterogeneity of coordinate expressions and unbalanced precision formats presents a considerable challenge for EM accuracy.

\begin{table}[htb]
\resizebox{\columnwidth}{!}{
\begin{tabular}{rcccccc}
    & \multicolumn{2}{c}{D} & \multicolumn{2}{c}{DM} & \multicolumn{2}{c}{DMS} \\\hline
        & mt5       & gpt       & mt5       & gpt       & mt5       & gpt   \\ \hline
D       & 0.81      & 0.81      & \textbf{0.84}      & 0.77      & -         & -     \\
DM      & \textbf{0.92}      & 0.87      & \textbf{0.98}      & 0.91      & 0.5       & \textbf{0.83}  \\
DMS     & 0.0       & 0.0       & 0.85      & 0.85      & 0.56      & \textbf{0.91}  \\ \hline          
\end{tabular}}
\caption{Exact match (EM) evaluation scores for coordinate retrieval, comparing MT5 and GPT model performance for coordinate precision levels. Rows represent latitude precision level. Columns represent longitude precision level.}
    \label{tab:extraction-scores-precision}
\vspace{-0.25cm}
\end{table}

\section{Out-of-Domain Experiments}

Next we apply these models to two additional sources: 1743 \textit{Dictionnaire universel françois et latin, dit de Trévoux} \citep{dictionnaire1743}, made available online by \citetlanguageresource{Trevoux1743}, and the 7th edition of the \citet{britannica1842} (1842), for which structured data has been shared by the \citetlanguageresource{EB2025}. 
From Trévoux, we identified 13,764 ``Geography'' articles using a fine-tuned classification model trained on the \textit{Encyclopédie} \citep{brenon2022classifying}.
420 of these contain coordinates according to our classification model (see Section~\ref{sec:classification}). Manually reviewing the 100 first entries shows that 88 contain coordinates and 61\% of those are EMs.
From 21,118 \textit{Britannica} entries, we process a random sample of 1,000 with our binary classification model, identifying 179 articles with coordinates. Manual validation confirms 172 of these as true positives (96\%). Then we apply the \texttt{mt5-small} normalization model.
Manual inspection showed 133 EMs (77\%), with most errors occurring in entries containing surface coordinates, lengthy articles, or coordinates expressed in textual, rather than numeric, form. Although these results are promising, deeper analysis of these datasets lies beyond the scope of this study.

\section{Related Work}
The dominant focus in automatically identifying spatial information in text data---both modern and historical---has been on recognizing place names, and linking and resolving these to knowledge base or gazetteer records \citep{jones2008, mcdonough2019, ehrmann2023}. Other forms of spatial information have been understudied, including the coordinates we study here. Methods for coordinate retrieval or extraction has, to date, focused on modern, usually scientific literature \citep{acheson2021}. 
As part of broader investigations of geography in French encyclopedias \citep{vigier2022}, \citet{moncla2024geoedda} automatically retrieve coordinates from \textit{historical texts} for the first time, and, here, we add the subtask of \textit{normalizing} these coordinates so that they are usable in a digital context 

Historical reference texts contain evidence about the development of scientific writing, including geographical discourse. As more encyclopedia and other major geographical text collections are shared as encoded data, it becomes easier to explore patterns in geographic information and description at scale. The Text Encoding Initiative guidelines \citeyearpar{tei} reflect best practices in annotating elements like pages, articles, and paragraphs. For example, beyond the two digital editions of the \textit{Encyclopédie} from ARTFL and ENCCRE, \textit{Britannica} data is available from the Nineteenth-Century Knowledge Project and the National Library of Scotland\footnote{\url{https://data.nls.uk/data/digitised-collections/encyclopaedia-britannica}}, and \citet{hagen-etal-2020-twenty} have released 22 German encyclopedias.

The coordinate annotation described here is related to sequence named entity annotation, specifically for parts of speech and named entities. Thanks to shared tasks, such annotations have been extremely popular: for example, \citet{tjong-kim-sang-buchholz-2000-introduction} on nominal chunks, \citet{tjong-kim-sang-dejean-2001-introduction} on clauses, \citet{tjong-kim-sang-2002-introduction} and \citet{tjong-kim-sang-de-meulder-2003-introduction} on named entity recognition. 

Sequence annotation methods include support vector machines, feed-forward neural networks, long short-term memory networks, and transformer encoders. Milestones include \citet{Kudoh2000}, \citet{Collobert2011}, \citet{Sutton2011}, \citet{hochreiter1997long}, and \citet{Devlin2019}. \citet{6978954} and \citet{soil-9-155-2023} are dedicated to the extraction of geographic information and coordinates.
Here, we use transformers and sequence-to-sequence models to extract coordinates \citep{Vaswani2017}. While transformers, in the form encoder-decoder \citep{raffel2020exploring,xue-etal-2021-mt5}, encoders \citep{Devlin2019}, or decoders \citep{liu2023empirical,wang2023gpt} have been applied to information extraction generally, to the best of our knowledge, none has been applied to coordinate retrieval or extraction.

\section{Conclusion}

Our results illustrate the strengths and limitations of transformer-based models in historical coordinate retrieval and normalization. While they excel at capturing regular patterns in notation, as evidenced by the very low character-level CER scores, they still face difficulties in normalizing commonly inconsistent historical formats. This is particularly true for the DMS format, where even minor errors can significantly impact EM performance.

To address these challenges, future research could focus on integrating rule-based post-processing techniques to normalize common formatting inconsistencies. Additionally, hybrid approaches that combine neural extraction with symbolic rules may offer a pathway to improved EM performance. 
Another promising direction would be to compare the distance between the coordinates documented in the text (locating a place) and modern coordinates (locating the same place). In other words, historical coordinates are not errors  because they are not perfect matches for modern coordinates: there are many reasons why there might be a difference, and this interpretation will be the subject of future research. To pursue this research, we will use \textit{Encyclopédie} place names linked to Wikidata items \citep{nugues:2024:LREC}.
By advancing these methods, we can enhance non-standard, historical coordinate retrieval and nomalization. New research would thus open doors to analyzing both the diversity of how coordinates are expressed in texts and the spatial distribution of coordinates around the earth. Without the ability to transform unstructured, non-standard coordinates in texts to accepted forms of digital geospatial data, our ability to explore coordinates as spatial data will remain limited. The gold standard data, fine-tuned models, and demonstrations presented here are a first step towards enabling historical spatial data analysis based on coordinates, not just named places.%

\section{Ethics Statement and Limitations}
The original text of Diderot and d'Alembert's \textit{Encyclopédie} is in the public domain. We use digitized versions provided by ENCCRE\footnote{\url{https://enccre.academie-sciences.fr/encyclopedie/}} \citeplanguageresource{ENCCRE2017} and ARTFL\footnote{\url{https://encyclopedie.uchicago.edu/}} \citeplanguageresource{ARTFL2022}. This encyclopedia contains geographical information that is sometimes obsolete and possibly false. In particular, the coordinates we have retrieved do not follow the standard format used today and our normalized coordinates may contain errors in comparison to modern locations. The geographical entries describing some regions or people may contain historical prejudices.

Although the automatic binary classification on the presence or absence of coordinates in an entry is nearly perfect, this is not the case for the retrieval and normalization of the coordinates. Further research on models and training methods is needed to improve on our results. We conduct out-of-domain experiments on small datasets: additional manual, expert annotation would be required for a full-scale evaluation of these models on the same tasks across multiple historical text genres and languages.

\section{Acknowledgements}
The authors are grateful to the ASLAN project (ANR-10-LABX-0081) of the Université de Lyon, for its financial support within the French program "Investments for the Future" operated by the National Research Agency (ANR). 
This work was partially supported by \textit{Vetenskaprådet}, the Swedish Research Council, registration number 2021-04533.

\section{Bibliographical References}\label{sec:reference}

\bibliographystyle{lrec2026-natbib}
\bibliography{bibliographie}

\section{Language Resource References}
\label{lr:ref}
\bibliographystylelanguageresource{lrec2026-natbib}
\bibliographylanguageresource{languageresource}

\end{document}